\pdfoutput=1
\PassOptionsToPackage{dvipsnames,svgnames,table}{xcolor}

\documentclass[11pt]{article}

\usepackage{acl}

\usepackage{times}
\usepackage{latexsym}
\usepackage{multirow}
\usepackage{makecell}

\usepackage[T1]{fontenc}

\usepackage[utf8]{inputenc}

\usepackage{microtype}

\usepackage{inconsolata}

\usepackage{kotex}

\usepackage{multirow}
\usepackage{multicol}
\usepackage{enumitem}
\usepackage{array}
\usepackage{amsmath}
\usepackage{graphicx}
\usepackage{cleveref}

\usepackage{xcolor}

\usepackage{hyperref}

\usepackage{array} 

\usepackage{setspace}

\title{Which is better? Exploring Prompting Strategy For LLM-based Metrics}

\author{Joonghoon Kim \qquad Saeran Park \qquad Kiyoon Jeong  \\
\textbf{Sangmin Lee} \qquad \textbf{Seung Hun Han} \qquad \textbf{Jiyoon Lee} \qquad \textbf{Pilsung Kang\text{*}} \\ \\
Korea University, Seoul, Republic of Korea\\ 
\normalsize\texttt{\{joonghoon\_kim,saeran\_park,kiyoon\_jeong,sangmin\_lee,andrewhan,jiyoon\_lee,pilsung\_kang\}}\\ \normalsize\texttt{@korea.ac.kr}}

\begin{document}
\maketitle
\begin{abstract}
This paper describes the DSBA submissions to the Prompting Large Language Models as Explainable Metrics shared task, where systems were submitted to two tracks: small and large summarization tracks. 
With advanced Large Language Models (LLMs) such as GPT-4, evaluating the quality of Natural Language Generation (NLG) has become increasingly paramount. 
Traditional similarity-based metrics such as BLEU and ROUGE have shown to misalign with human evaluation and are ill-suited for open-ended generation tasks. 
To address this issue, we explore the potential capability of LLM-based metrics, especially leveraging open-source LLMs.
In this study, wide range of prompts and prompting techniques are systematically analyzed with three approaches: prompting strategy, score aggregation, and explainability. 
Our research focuses on formulating effective prompt templates, determining the granularity of NLG quality scores and assessing the impact of in-context examples on LLM-based evaluation. Furthermore, three aggregation strategies are compared to identify the most reliable method for aggregating NLG quality scores.
To examine explainability, we devise a strategy that generates rationales for the scores and analyzes the characteristics of the explanation produced by the open-source LLMs.
Extensive experiments provide insights regarding evaluation capabilities of open-source LLMs and suggest effective prompting strategies.\footnote{Code for this paper is available at \url{https://github.com/kjhoon7686/Prompt4LLM-Eval}.}
\end{abstract}

\section{Introduction}
As Large Language Models (LLMs) like GPT-4 continue to advance rapidly, the Natural Language Generation (NLG) capability is approaching a level of expertise comparable to that of a human. As a result, the precise evaluation of NLG has become increasingly paramount. 
However, traditional similarity-based metrics like BLEU \citep{b2} and ROUGE \citep{b3}, which are widely used in NLG evaluations, tend to show a discrepancy from human assessments \citep{b1}. 
Additionally, the reliance on reference texts for these metrics can hinder an accurate assessment of NLG quality, particularly for open-ended generation tasks.

Recent research has introduced methodologies that leverage LLMs as NLG evaluators, showcasing the potential of LLM-based metrics. 
These approaches are motivated from findings in recent research which revealed that LLM can directly evaluate NLG capabiltiy harnessing knowledge retained during the pre-train \citep{b14}.
These metrics have demonstrated notable correlation \cite{b16, b1, b19-gemba, b20-automqm} with human evaluations to learned evaluators \citep{b21-canLLM, b29-learned_evaluator}.

Concurrently, recent advancement of LLMs such as LLaMA \citep{b8}, Vicuna \citep{b33-vicuna}, and Orca \citep{b12}, has paved a way for research on NLG evaluations utilizing open-source LLMs \citep{b14}.
However, there are few comprehensive studies that systematically evaluate the vast amount of possible prompts and prompting techniques for LLM-based metrics. 
Especially, research assessing the capabilities of open-source LLMs in the context of LLM-based metrics is even more scarce. 
Given the importance of enhancing the reproducibility of LLM-based metrics in metric research, there is a clear need for studies that explore effective prompts and prompting techniques specifically for open-source LLMs \cite{b21-canLLM}.

In this work, we conduct a thorough exploration of various prompts and prompting techniques for effective deployment of open-source LLMs as metrics: analyze them in terms of prompting strategy, score aggregation, and explainability.

Within the scope of prompting strategies, we compare the effectiveness of human and model instruction templates for NLG evaluation. In addition, we explore granularity in score assignment to accurately evaluate NLG quality. 
Additionally, we gauge the influence of the open-source LLM's In-Context Learning (ICL) capability \cite{b22-ICL} in NLG evaluation by employing various types of demonstrated examples.
For score aggregation, we compare three methodologies to discern the optimal strategy for aggregating NLG quality scores.
To infer the explainability of open-source LLMs, we generate rationale when computing scores. These comprehensive experiments on prompting techniques for LLM-based metrics provide insights into the evaluation capabilities of open-source LLMs and guidelines for effective prompting strategies.

Furthermore, we provide insights derived from analysis of the features embedded in prompts and behaviors of open-source LLMs as LLM-based metrics. 
Additionally, we report our strategies and outcomes applied to the test set of summarization track in Eval4NLP 2023 shared task.

\begin{figure*}[!htp]
    \centering
    \includegraphics[width=0.75\linewidth]{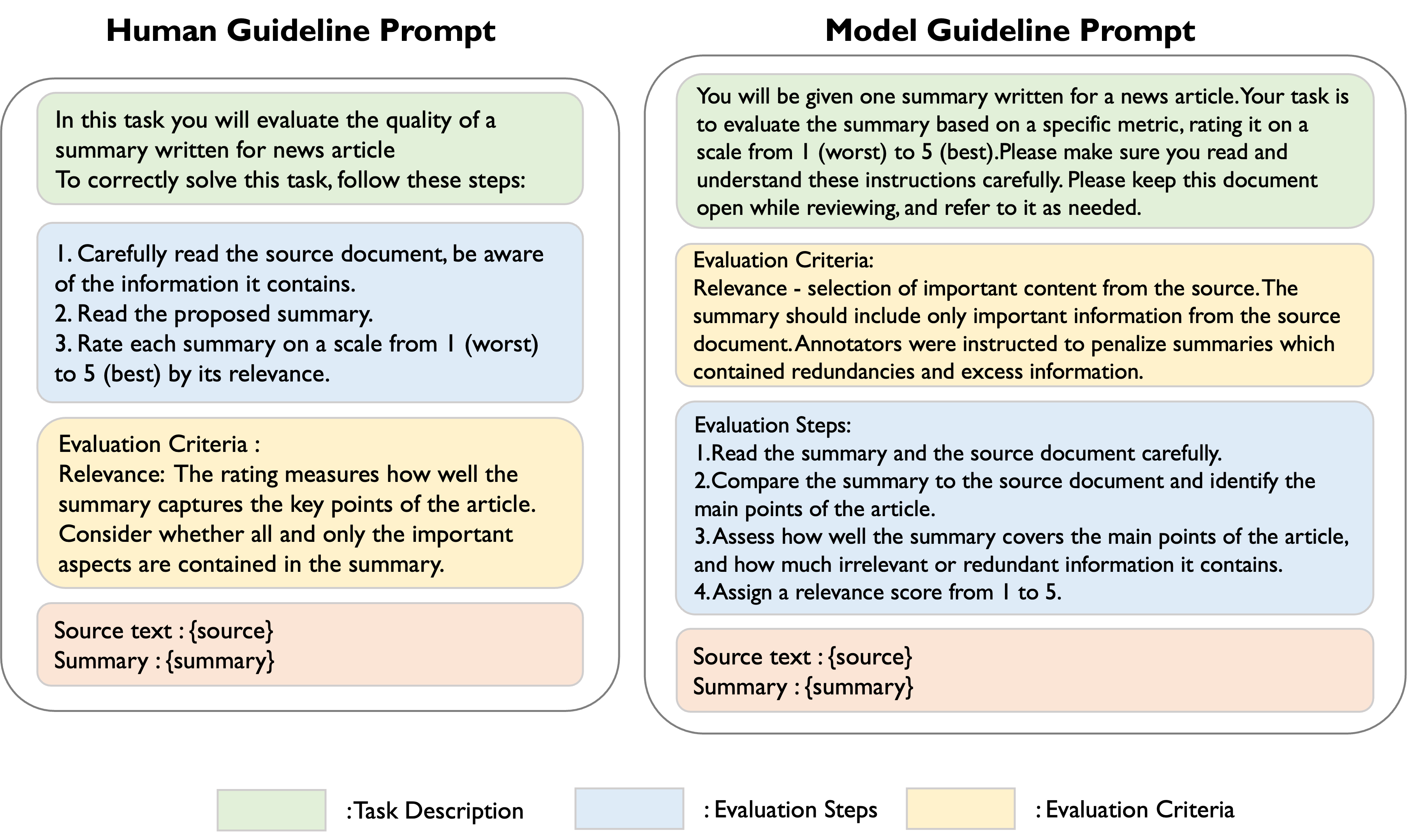}
    \caption{Examples of Human Guideline (HG) prompt and Model Guideline (MG) prompt. 
    HG prompt and MG prompt consists of task description, evaluation criteria, and evaluation steps. 
    The HG prompt is used as the annotation guideline for summarization evaluation, serving as the basis for human annotators assessments. 
    In contrast, the MG prompt was used as the instruction for the model.}
    \label{fig:1}
\end{figure*}

\section{Related Work}
\textbf{Similarity-based Metrics}
\quad Similarity-based metrics evaluate the quality of NLG outputs by comparing reference and candidate text. 
They can be categorized into lexical-based and semantic-based metrics.
Lexical-based metrics, such as BLEU \cite{b2} and ROUGE \cite{b3}, utilize N-grams to measure lexical overlap between a reference and a candidate text. 
However, research has highlighted their inadequacy in accurately assessing the quality of generated outputs and identifying both syntactical and semantic discrepancies \cite{b1, b25-limitNgram, b26-limitNgram2}. 
On the other hand, semantic-based metrics, including BERTScore \cite{b4} and MoverScore \cite{b5}, measure semantic similarity by comparing the embeddings of both reference and candidate texts. 
However, similar to lexical-based metrics, they face challenges when evaluating open-ended generation tasks due to their inherent dependence on reference text \cite{b21-canLLM, b27-openmeva, b28-perceptionscore}. 
\\
 
\noindent \textbf{LLM-based Metrics} 
\quad The recent substantial advancement in the NLG capabilities of LLMs has motivated research interests related to LLM-based metrics. Consequently, the latest studies, primarily exploring various prompting approaches that do not require additional training of an LLM, has shown a correlation with human evaluation comparable to that of learned evaluators \cite{b21-canLLM, b29-learned_evaluator}. 
Also, building upon the foundational work of LLaMA \cite{b8}, research on the fine-tuning approach which constructs an evaluator by fine-tuning an LLM with suitable supervised data for the evaluation task, is being actively pursued \cite{b13, b14}.

\section{Summarization Track}
\label{section:Summarization Track}
The summarization track of Eval4NLP 2023 shared task \cite{eval4nlp23} aims to propose a reference-free metric for summarization.
Specifically, reference-free metric evaluates a given summary using only the provided source sentence or paragraph without additional human-written references. 
The objective of shared task is to develop LLM-based metrics by exploring effective prompting strategies for open-source LLMs. 


\subsection{Dataset}
\label{section:dataset}
\subsubsection{Train and Development Set}
In this study, we utilize the SummEval benchmark dataset provided by \citet{b15} as both train and development sets. 
While the original benchmark provides human annotation scores for each of four aspects, including \texttt{relevance}, \texttt{consistency}, \texttt{coherence}, and \texttt{fluency}, the summarization track adopts the average of these aspect scores as golden human annotation scores. 
The performance of the evaluation task is measured through sentence-level correlation with the golden human annotation scores.

\subsubsection{Test Set}
Dataset provided in the shared task \cite{eval4nlp23}, consisting of sentences and fragments of paragraphs from English Wikipedia documents written after July 15, 2023, is used as the test set. Summaries in the test dataset were generated by a summary generation model that are annotated with reference to Multidimensional Quality Metrics (MQM) annotation for aspects like \texttt{factuality}, \texttt{relevance}, and \texttt{readability}.

\subsection{Models}
\label{section:model}
We use four out of six open-source LLMs provided in the Eval4NLP 2023 shared task.

\begin{itemize}[noitemsep]
\item \textbf{Hermes-13B} - LLaMA-13B model trained on over 300,000 instructions.
\item \textbf{Orca-7B} - LLaMA2-7B model trained on Orca Style dataset.
\item \textbf{Orca-13B} - LLaMA2-13B model trained on Open-Platypus dataset and OpenOrca dataset.
\item \textbf{Platypus-70B} - LLaMA2-70B model trained by \citet{b10}.
\end{itemize}

\section{Method}
In this section, we address the prompting strategies and score aggregation methods, as well as approaches to assess the explainability of open-source LLMs.

\subsection{Prompting Strategy}
\label{prompt_strategy}
Prompting strategies consist of prompt template, granularity of score, and demonstration.


\subsubsection{Prompt Template}
We propose Human Guideline (HG) prompt and Model Guideline (MG) prompt for summary evaluation as illustrated in Figure \ref{fig:1}. 
The HG prompt, adapted from the human evaluation guideline of SummEval \cite{b15}, provides clear evaluation instructions and criteria for human annotators. 

Conversely, the MG prompt, implemented from a guideline given to LLM such as GPT-4 for summary evaluation in G-EVAL \cite{b1}, instructs LLM to assess summaries, offering detailed, directive instructions and criteria.

Both HG prompt and MG prompt consist of elements such as task description, evaluation criteria, and evaluation steps.
To assess the impact of each element, we create variants by modifying each one.
\\

\noindent \textbf{Task Description} 
\quad The task description provides instructions for the specified task. 
To explore the influence of its length, we craft short and long descriptions by varying sentence lengths, maintaining the original context.
Additionally, we create an expert-role task description to study the effect of providing an expert role in the evaluation (e.g. ``you're an expert at summarizing news articles."). 
Each variant is developed for both HG and MG prompts, with details in Appendix \ref{appendix:D}.
\\

\noindent \textbf{Evaluation Criteria} 
\quad The evaluation criteria outlines the scoring standards for the given summary per aspect. 
It is categorized into three components, 1) Aspect Definition (AD) 2) Human-Targeted criteria (HT) 3) Model-Targeted criteria (MT).

AD, adopted from GPTScore \cite{b16}, concisely describes the evaluation aspect definitions. HT and MT, used in HG and MG Prompts respectively, include scoring considerations and aspect descriptions. 

To investigate the effects of each components, we generate modified version of AD, HT, and MT for each aspect using GPT-4.
We instruct GPT-4 to maintain a consistent format with the existing ones.
Examples are provided in Appendix \ref{appendix:D}.
\\


\noindent \textbf{Evaluation Steps} 
\quad The evaluation steps, which could be considered as a Chain-of-Thought (CoT) \cite{b17}, provide step-by-step instructions for the evaluation task, enhancing the reasoning capabilities of LLM. 
To explore the impact of varied evaluation steps descriptions, we construct detailed complex evaluation steps for both HG and MG prompts. Examples are provided in Appendix \ref{appendix:D}.

\subsubsection{Granularity of Score}
For assigning a score, we consider the following two scoring approaches: coarse-grained scoring and fine-grained scoring. Coarse-grained scoring yields a singular and holistic score that considers all evaluation aspects collectively, but does not provide scores for individual aspects.
Conversely, fine-grained scoring assigns the score for each aspect, deriving individual scores and then averaging them to yield the final singular score. 
This approach enables the LLMs to furnish both the overall score and specific aspect scores, granting a more nuanced understanding of for score derivation compared to the coarse-grained method. 
Given that NLG evaluations commonly score by jointly taking multiple aspects into account, adpoting fined-grained scoring when constructing variants of the prompt is naturally apt approach.

\subsubsection{Demonstration}
\label{demo}
To examine the ICL capability of open-source LLMs in evaluation tasks, we craft two distinct types of demonstrated examples.

One set of examples includes raw source text, a summary, and a human annotation score.
On the other hand, another set of examples incorporates a rationale derived from the assigned human annotation score, which has been distilled from GPT-4\footnote{https://openai.com/research/gpt-4}, in addition to the components found in the former set of examples.
Examples are provided in Appendix \ref{appendix:D}.

Furthermore, we construct examples for each individual aspect and subsequently group them into 'worst' and 'best' categories based on human annotation scores. 
In our study, 'worst' examples are assigned a score of 1, while 'best' examples receiving a score of 5.
Categorization is undertaken to investigate potential biases in the quality and the score of the provided examples.
Due to the maximum input length constraint of the LLMs, we use only one example as demonstration per summary.


\begin{figure*}[!htp]
    \centering
    \includegraphics[width=0.7\linewidth]{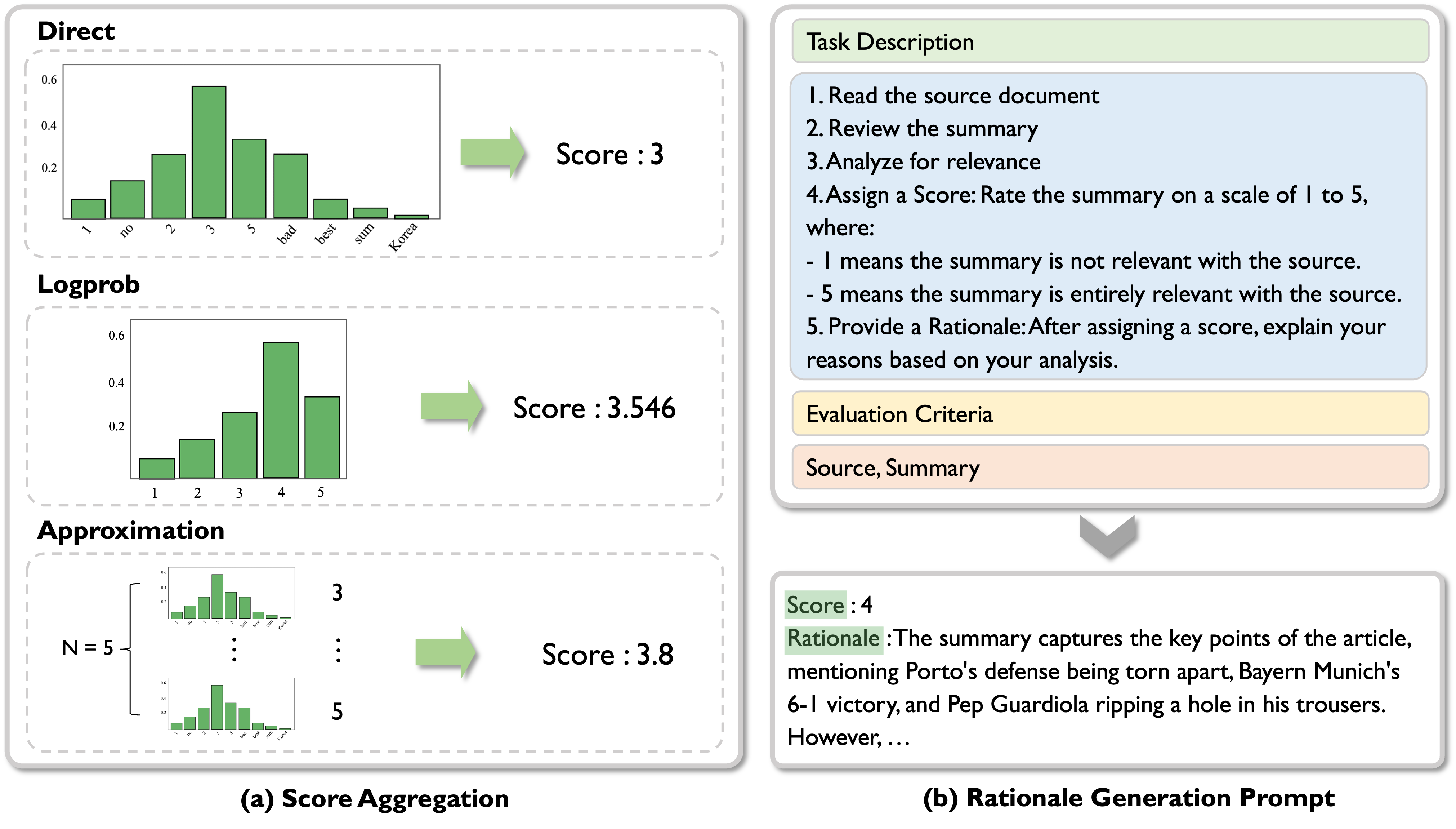}
    \caption{\textbf{(a) Left - Score Aggregation} An example of how the Score Aggregation is calculated. 
    `Direct' uses scores directly generated by the model, `Logprob' uses a weighted summation based on generation probabilities of pre-defined scores (e.g. 1 to 5), and `Approximation' uses an average from N sampled scores. \textbf{(b) Right - Rational Generation prompt} An example of Rationale Generation (RG) prompt and the corresponding outputs.
    Using the RG prompt as input, the model provides a score for the quality of the summary and the corresponding rationale.}
    \label{fig:2}
\end{figure*}
\subsection{Score Aggregation}

To derive scores for individual aspects, we propose the following three score aggregation methods: Direct, Logprob, and Approximation (see Figure \ref{fig:2}).
\\

\noindent \textbf{Direct} 
\quad  This method is the most general scoring method. It leverages the score generated by the LLM directly.
\\

\noindent \textbf{Logprob} 
\quad  This method calculates the score by summing the product of a pre-defined discrete score range (e.g. 1 to 5) and the generation probability of the corresponding tokens. 
This method is considered as a weighted summation approach, using each score's token probability as its weight.
By incorporating the model's token generation probabilities, this method distinctively produces a more continuous score.

For a given set of pre-defined discrete scores \(S = \{{s_1,...,s_K}\}\), Logprob multiplies each discrete score \( s_i \) by its token 
probability \( p(s_i) \). 
K in \eqref{eq1} is the number of pre-defined discrete scores.

\begin{equation} 
score = \sum_{i = 1}^K  p(s_i) \cdot s_i 
\label{eq1}
\end{equation}

\noindent \textbf{Approximation} 
\quad  This method calculates the score by averaging N sampled scores generated by LLM. 
Intending to approximate the token probability distribution, we design Approximation method to distinguish it from the Logprob method, which directly uses the actual token probabilities. 
This aggregation is inspired by techniques explored in \citep{b1, b16}.

For a given set of pre-defined discrete scores \(S = \{{s_1,...,s_K}\}\), Approximation multiplies each discrete score \( s_i \) by its approximated token probability \( g(s_i) \). In (\ref{eq2}), \(count(s_i\)) denotes the number of count discrete score \(s_i\) appears in N samples.

\begin{align}
g(s_i) &= \frac{{count(s_i)}}{N} \label{eq2} \\
\text{{score}} &= \sum_{i=1}^{K} g(s_i) \cdot s_i \label{eq3}
\end{align}

\subsection{Explainability}
\label{explainability}
Evaluations that employ the previously described methods yield only a sole scalar score with no additional explanation for the assigned score at all. 
Thus, we manually craft the Rationale Generation (RG) prompt to derive rationales for the scores. 
Using this prompt, we aim to explore the explainability of open-source LLMs (see Figure \ref{fig:2}).

Furthermore, similar to the approach used in the demonstration section \ref{demo}, we use examples to analyze the influence of demonstrated examples on rationale generation.
Each example is divided into `worst' and `best' example to examine potential biases in the outputs.

\subsection{Test phase}
For the test set, we incorporate two supplementary approaches alongside the previously described prompting strategy, tailored to the attributes of the test set.
\\

\noindent \textbf{Filtering} 
\quad  Although many summaries in the test set exhibit appropriate sentence structures, certain samples retain repetitive words or phrases (e.g. ``A family of four members, including a first member, a second member, a third member, and a fourth member.").
We deem such instance as a failure to generate an appropriate summary and uniformly assigned them lowest score. 
To account such instances, we design a Filtering prompt that filters failed samples.
For given summaries, when model generates a `Yes' response, they are assigned the minimum score. 
Example of the Filtering prompt is provided in Appendix \ref{appendix:D}.
\\

\noindent \textbf{Binning} 
\quad  After analyzing the scores assigned by the model for the test data, we observe that open-source LLMs are generally adept at evaluating summaries. 
Nevertheless, we note the model's tendency of assigning excessively fine-grained scores among samples of equivalent quality (e.g. scores of 1 and 1.01). 
In light of these observations, we implement Binning to simplify the score distribution and mitigate noise, thereby integrating proximate scores into same categories.
Detailed explanations can be found in the Appendix \ref{appendix:B}.

\begin{table*}[!ht]
\centering
\resizebox{300pt}{!}{%
\begin{tabular}{cccc|cc}
\Xhline{1pt}
\textbf{Template}   & \textbf{Fine-grained} & \textbf{Demonstration} & \textbf{Aggregation}  & \textbf{Orca-7B}                         & \textbf{Orca-13B}     \\ \hline\hline
\textbf{\textrm{Prompting}}         & \multicolumn{1}{l}{}  & \multicolumn{1}{l}{}   & \multicolumn{1}{l|}{} & \multicolumn{1}{l}{}                    & \multicolumn{1}{l}{} \\
Base                       & x                     & x                      & Direct                & 0.2500                                    & 0.3040                \\
Human                      & x                     & x                      & Direct                & 0.3094                                  & 0.4343               \\
Model                      & x                     & x                      & Direct                & 0.2651                                  & 0.3583               \\
\rowcolor[HTML]{EFEFEF} 
Base                       & o                     & x                      & Direct                & 0.2746                                  & 0.3891               \\
\rowcolor[HTML]{EFEFEF} 
Human                      & o                     & x                      & Direct                & \textbf{0.3472}                         & \textbf{0.4468}      \\
\rowcolor[HTML]{EFEFEF} 
Model                      & o                     & x                      & Direct                & 0.2864                                  & 0.3844               \\ \hline
\textbf{\textrm{Demonstration}}     & \multicolumn{1}{l}{}  & \multicolumn{1}{l}{}   & \multicolumn{1}{l|}{} & \multicolumn{1}{l}{}                    & \multicolumn{1}{l}{} \\
Human                      & o                     & Base-worst             & Direct                & 0.1758                                  & 0.3690                \\
Human                      & o                     & Base-best              & Direct                & \textbf{0.2854}                         & 0.4092               \\
\rowcolor[HTML]{EFEFEF} 
Human                      & o                     & Reason-worst           & Direct                & 0.2309                                  & 0.3899               \\
\rowcolor[HTML]{EFEFEF} 
Human                      & o                     & Reason-best            & Direct                & 0.2733                                  & \textbf{0.4133}      \\ \hline
\textbf{\textrm{Aggregation}} & \multicolumn{1}{l}{}  & \multicolumn{1}{l}{}   & \multicolumn{1}{l|}{} & \multicolumn{1}{l}{}                    & \multicolumn{1}{l}{} \\
Human                      & o                     & x                      & Approximation         & 0.3239                                  & 0.4002               \\
Human                      & o                     & x                      & Logprob               & 0.3296                                  & 0.4210                \\
Human                      & o                     & x                      & Direct                & \textbf{0.3472}                         & \textbf{0.4468}      \\
\rowcolor[HTML]{EFEFEF} 
Model                      & o                     & x                      & Approximation         & 0.2687                                  & 0.3530                \\
\rowcolor[HTML]{EFEFEF} 
Model                      & o                     & x                      & Logprob               & \textbf{0.2926}                         & \textbf{0.3851}      \\
\rowcolor[HTML]{EFEFEF} 
Model                      & o                     & x                      & Direct                & 0.2864                                  & 0.3844               \\ \hline
\textbf{\textrm{Explainability}}    & \multicolumn{1}{l}{}  & \multicolumn{1}{l}{}   & \multicolumn{1}{l|}{} & \multicolumn{1}{l}{}                    & \multicolumn{1}{l}{} \\
Rationale                  & o                     & x                      & Direct                & \cellcolor[HTML]{FFFFFF}\textbf{0.3506} & 0.4220                \\
\rowcolor[HTML]{EFEFEF} 
Rationale                  & o                     & Reason-worst           & Direct                & 0.2915                                  & 0.3876               \\
\rowcolor[HTML]{EFEFEF} 
Rationale                  & o                     & Reason-best            & Direct                & 0.3262                                  & \textbf{0.4330}       \\ \Xhline{1pt}
\end{tabular}%
}
\caption{Main result. Experimental results of combination sets for each Prompting Strategy, Score Aggregation, and Explainability. `Human' and `Model' mean Human Guideline prompt and Model Guideline prompt respectively. Also, `Base-worst/best' and `Reason-worst/best' are abbreviations of two types of demonstration that are distinguished, including rationale. Best results for each set of variants are in bold.}
\label{tab:tab_1}
\end{table*}

\section{Experiments}
\subsection{Experimental Setup}
Experiments are conducted using the development set of the summarization track provided in the shared task.
We use the provided prompt template for the summarization track as the baseline prompt. 
The baseline prompt contains a brief task description and score guide. 
Additionally, the HG and MG prompt in \ref{main results} are adapted from SummEval \cite{b15} and G-EVAL \cite{b1} with minimal modification.
Examples of prompts are provided in Appendix \ref{appendix:D}. 
For scoring, we averaged the scores derived from the aspects of \texttt{relevance}, \texttt{consistency}, \texttt{coherence}, and \texttt{fluency} for fine-grained scoring.
For the demonstration experiments, we sample examples from the train set based on human annotation scores for each aspect. 
Rationales for the scores in the examples are generated using GPT-4.
Throughout the entire score generation process, we set top\_p to 0.1. 
For Direct and Logprob aggregation, the temperature is set to 0. 
Lastly, we set the temperature to 1 and n\_samples to 20, respectively, for Approximation aggregation. 

Moreover, we report the leaderboard results for the test set using Orca-13B and Platypus-70B for the small and large track, respectively.
Test set experiments share the almost the same setting with development set experiments: same HG prompt, fine-grained scoring, hyperparameters for Direct aggregation are implemented. 
For \texttt{factuality} evaluation criteria, not originally provided in SummEval \cite{b15}, we use GPT-4 to generate it. 
Specifically, scores for \texttt{relevance}, \texttt{factuality}, and \texttt{fluency}, obtained from Direct aggregation, are averaged to compute the final score. 
Throughout our all experiments, segment-level Kendall's Tau correlation is used as the performance metric.
For optimized inference with open-source LLMs, we employ Guidance\footnote{https://github.com/guidance-ai/guidance} and vLLM\footnote{https://github.com/vllm-project/vllm} libraries. 
Details of experimental setup are provided in Appendix \ref{appendix:A}.

\subsection{Main Results}
\label{main results}

\subsubsection{Prompting Strategy}

We compare the performance with different types of the prompt templates.
As shown in \textbf{Prompting} section of Table~\ref{tab:tab_1}, regardless of the granularity of the score, we observe that HG and MG prompts, especially HG prompt, consistently outperform the baseline prompt.
We hypothesize that a more detailed description of task provided in the HG and MG prompt allows LLM to understand and follow the instructions more clearly. 
Moreover, among all the prompts, the HG prompt achieves the best performance, indicating that succinct and clear instructions are better than complex ones.

As for granularity of the scoring, fine-grained scoring consistently outperforms coarse-grained scoring across various model sizes and prompt templates. 
The coarse-grained scoring may introduce ambiguity in the evaluation criteria by requiring the LLM to consider aspect-specific considerations in an integrated manner. 
Conversely, the fine-grained scoring removes such ambiguity by providing evaluation criteria of each aspect independently.

As shown in \textbf{Demonstration} section of Table~\ref{tab:tab_1}, we observe that the use of demonstration leads to decrease in performance, likely due to the inherent bias introduced by the demonstrated example.
Notably, the smaller model exhibits a significant decline in performance, which could be attributed to their limited ICL capabilities \citep{b30-icl_survey, b31-icl2, b32-icl3}, resulting in inaccurate understanding of in-context examples, and vice versa. 
The performance differs among models based on whether they are provided with examples containing only the score or examples with additional rationales. 
This discrepancy can be attributed to the superior ability of larger models in comprehending in-context examples, which leads to better understanding when explanations for scores are added. 
In contrast, the smaller model exhibits the opposite behavior.
Furthermore, providing the `best' examples consistently yields superior performance across all model sizes when compared to the `worst' examples. 
After conducting an analysis of the model's score distribution, we observe a bias wherein the model tends to assign higher scores when provided with the `best' example.
We hypothesize that observed bias may be driven by the skewed distribution of human annotation scores in the development set, where human annotation scores are predominantly distributed towards higher values, mainly falling between 3 and 5.

\subsubsection{Score Aggregation}
We assess the performance based on the different score aggregation methods.
\textbf{Aggregation} section of Table~\ref{tab:tab_1} illustrates that, across various model sizes and prompt templates, Direct and Logprob aggregation consistently demonstrates superior performance when compared to the Approximation aggregation.
In both Direct and Logprob aggregation, the decoding temperature is set to 0. 
This likely leads the model to assign scores in a more deterministic manner compared to the Approximation, potentially resulting in superior performance. 
Specifically, since Approximation estimates the distribution of score token probability through sampling, sampling noise could account for its lower performance.
Unlike other aggregation methods, Direct aggregation generates integer values ranging from 1 to 5, thereby offering a much fewer score range.
On the other hand, \citet{b14} suggest that Kendall Tau might favor tie pairs.
Such tendency could explain the notably high correlation observed with Direct aggregation. 



\subsubsection{Explainability}
We assess the LLM's ability to provide appropriate explanations for the scores.
Examining \textbf{Explainability} section of Table~\ref{tab:tab_1}, we observe that the RG prompt results in performance similar to or slightly lower than the HG prompt and better than the MG prompt. 
This suggests that generating rationales for scores can also aid the evaluation process itself.
Furthermore, it is noteworthy that Orca-7B exhibits a slight performance decline when provided with a demonstrated example, in contrast to the performance of Orca-13B.
The RG prompt is meticulously designed to facilitate the generation of rationales, possibly benefiting from the examples.
Therefore, Orca-13B, with superior ICL capabilities as mentioned in \ref{prompt_strategy}, has outperformed the other smaller model.
Analysis of the rationales generated by Orca-13B is discussed in \ref{err_analysis}.


\subsubsection{Test Phase}

\begin{table}[!ht]
\centering
\resizebox{180pt}{!}{%
\begin{tabular}{ccc}
\Xhline{1pt}
               & \textbf{Orca-13B} & \textbf{Platypus-70B} \\ \hline\hline
Human           & 0.4699           & 0.4764               \\
Filtering     & 0.4815           & -                    \\
Binning & \textbf{0.5016}           & 0.4916               \\ \Xhline{1pt}
\end{tabular}%
}
\caption{Kendall's Tau correlation on test set where Human denotes test result obtained with HG prompt.}
\label{tab:tab_2}
\end{table}

In Table~\ref{tab:tab_2}, we report the performance of the HG prompt on the test set.
Details of HG prompt applied for the test set are provided in Appendix \ref{appendix:D}.
As evident from the results of our development set experiments, the performance of the HG prompt on the test set is consistently satisfactory across all models.
Furthermore, we observe a discernible improvement in performance when the Filtering is applied.
This observation suggests that uniformly assigning lowest scores to inadequately generated summaries can enhance performance.
Similarly, Binning enhances performance by reducing noise in the scores on the test set. 
This improvement is achieved by integrating closely related scores into same categories.
While the Orca-13B model exhibits a slightly lower performance compared to the Platypus-70B with the base HG prompt, it shows superior performance after the application of Filtering and Binning. 
Details of test phase are provided in Appendix \ref{appendix:B}.
\subsection{Analysis}

\subsubsection{The Effect of Different Model Sizes}
We compare the performance depending on different model sizes: Orca-7B, Hermes-13B, Orca-13B, and Platypus-70B.
As shown in Appendix Table~\ref{tab:tab_b_1} and Table~\ref{tab:tab_b_2}, despite the same size with Orca-13B, the performance of Hermes-13B is significantly lower, even lower than Orca-7B. 
Except for Hermes-13B, generally positive correlation between model size and performance is observed. 
We speculate such outcome may be due to the differences in the backbone model's performance (e.g. LLaMA, LLaMA 2) and the type of datasets and approaches used for fine-tuning \cite{b18}. 
Insignificant performance gap between Platypus-70B and Orca-13B proves that Orca-13B is as effective as Platypus-70B for the evaluation task.

\subsubsection{Comparisons of each Component}

\noindent \textbf{Task Description Types} 
\quad We investigate the impact of varying the length of task descriptions within the HG prompt and MG prompt on performance. 
Additionally, we compare performance when an expert role is assigned in the task description versus when it is not.
As shown in Appendix Table~\ref{tab:tab_c_1}, for Orca-7B, there is no significant performance difference based on length of task descriptions. 
However, for Orca-13B, we observe higher performance when a longer task description is employed. 
Such tendency suggests that, Orca-13B benefits from longer length of task descriptions in facilitating the execution of instructions, even when the content remains the same.
Furthermore, when the expert role is assigned, there is a discernible performance improvement with Orca-7B. 
However, for Orca-13B, the performance difference between cases with and without the expert role is not substantial, indicating that this approach can be more effective for smaller models.
\\

\noindent \textbf{Evaluation Criteria Variants} 
\quad We analyze the influence of various evaluation criteria, AD, HT, and MT.
As shown in Appendix Table~\ref{tab:tab_c_2}, utilizing aspect definitions consistently improves performance, regardless of the prompt template or model size. 
Furthermore, similar results are obtained even when evaluation criteria generated by GPT-4 are used.
This suggests that providing a simple definition of each aspect is an effective approach when evaluating summary quality.
\\

\noindent \textbf{Complexity of Evaluation Steps} 
\quad As shown in Table~\ref{tab:tab_c_3}, there is no significant trend in performance between standard and complex evaluation steps both for the HG prompt and the MG prompt.
This observation implies that while the evaluation steps are effective in offering step-by-step instructions to the model, the precise description or complexity level of the evaluation steps does not exert a significant influence on the evaluation of summaries.

\subsubsection{Error Analysis} 
\label{err_analysis}
To investigate whether the model generates well-founded rationales for the assigned scores, we perform an error analysis on the rationales generated using the RG prompt described in section \ref{explainability}.
Specifically, we conduct such comparative analysis on 36 sampled instances for two different rationale generation method: one generated with Orca-13B and RG prompt, and another with RG prompt including demonstrated examples.

Our analysis reveals that, in general, the model exhibits the capability to provide rationales correctly. 
However, we identify several types of errors: \textbf{(Error type 1)} provided rationale is inconsistent with the assigned evaluation scores, \textbf{(Error type 2)} provided rationale shows hallucination where the rationale includes information not present in the source text or summary, \textbf{(Error type 3)} provided rationale describes explanation about aspect different from the designated one.
Detailed descriptions and examples for each error type can be found in Appendix \ref{appendix:C}.
Addressing and mitigating these errors through further research efforts could significantly enhance the explainability and reliability of LLM-based metrics.


\section{Conclusion}
In this work, we conduct a systematic analysis of effective prompting techniques and strategies for LLM-based metrics in NLG evaluation. 
Our comprehensive experiments reveal that providing clear and straightforward instructions, akin to those explained to humans, proves to be more effective.
Furthermore, we examine various score aggregation methods to achieve effective score assignments and show the potential for enhancing explainability within open-source LLMs. 
Additionally, we explore performance change relative to model size and scrutinize the influence of various elements within the prompt template.
We hope that our research findings will furnish valuable insights for future studies focused on LLM-based metrics, especially those leveraging open-source LLMs.

\bibliography{anthology,custom}

\appendix

\onecolumn\section{Experimental Setup}
\label{appendix:A}

\begin{table}[!ht]
\centering
\resizebox{0.2\textwidth}{!}{%
\begin{tabular}{c|c}
\Xhline{1pt}
\textbf{Library} & \textbf{Version} \\ \hline
guidance         & 0.0.64           \\
vllm             & 0.1.7            \\
torch          & 2.0.1            \\ \Xhline{1pt}
\end{tabular}%
}
\caption{Version of libraries used for the experiments.}
\label{tab:version}
\end{table}

For optimized inference with open-source LLMs, we employ Guidance and vLLM
libraries. 
The libraries and their respective versions used for the experiments can be found in Table~\ref{tab:version}.

\section{Test Phase}
\label{appendix:B}
We submit the final results for the test set after equally applying Filtering and Binning to the HG prompt on both Orca-13B and Platypus-70B (for the small and large track, respectively).
We use HT as the evaluation criteria of the \texttt{factuality}, generated using GPT-4.
Scores for \texttt{relevance}, \texttt{factuality}, and \texttt{fluency}, obtained from Direct aggregation, are averaged to compute the final score. 
The hyperparameters for Direct aggregation is set identical to the development set, with top\_p to 0.1 and temperature to 0, respectively.
The prompts used for the test set can be found in Table~\ref{tab:test_relevance},~\ref{tab:test_factuality}, and ~\ref{tab:test_fluency}.

Filtering is applied using the Filtering prompt on both Orca-13B and Platypus-70B models.
Example of the Filtering prompt is provided in Table~\ref{tab:appendix_d_6}.
After applying Binning, the number of unique scores has been diminished from 36 to 10 and 46 to 13 for Orca-13B and Platypus-70B, respectively.

\section{Analysis}
\label{appendix:C}
\subsection{The Effect of Different Model Sizes}
We conduct experiments to analyze the performance differences depending on model sizes using Orca-7B, Hermes-13B, Orca-13B, and Platypus-70B. 
The experiments for Orca-7B, Hermes-13B, and Orca-13B are conducted using vLLM, while the Platypus-70B experiments are conducted using Guidance. 
In Table~\ref{tab:tab_b_1}, we conduct experiments comparing performance across model sizes for different prompt templates and granularity of score. 
In Table~\ref{tab:tab_b_2}, we carry out experiments to compare performance across model sizes for different prompt templates and score aggregations.


\begin{table}[!ht]
\centering
\resizebox{\textwidth}{!}{%
\begin{tabular}{cccccccc}
\Xhline{1pt}
\textbf{Template} & \textbf{Fine-grained} & \textbf{Demonstration} & \textbf{Aggregation}       & \textbf{Orca-7B} & \textbf{Orca-13B} & \textbf{Hermes-13B} & \textbf{Platypus-70B} \\ \hline\hline
Base            & x            & x             & Direct  & 0.2500            & 0.3040            & 0.1554             & 0.3956               \\
Human           & x            & x             & Direct  & 0.3094          & 0.4343           & 0.2041             & 0.4260               \\
Model           & x            & x             & Direct  & 0.2651          & 0.3583           & 0.1915             & 0.4383               \\
\rowcolor[HTML]{EFEFEF}
Base            & o            & x             & Direct  & 0.2746          & 0.3891           & 0.1402             & 0.4082               \\
\rowcolor[HTML]{EFEFEF}
Human           & o            & x             & Direct  & 0.3472          & 0.4468           & 0.2063             & 0.4354               \\
\rowcolor[HTML]{EFEFEF}
Model           & o            & x             & Direct  & 0.2864          & 0.3744           & 0.2170             & 0.4039               \\ \Xhline{1pt}
\end{tabular}%
}
\caption{Comparison of Kendall's Tau correlation across various Prompt Templates and Models. \textit{Fine-grained} denotes whether the fine-grained scoring is used or not. \textit{Aggregation} denotes the type of Score Aggregation method used.}
\label{tab:tab_b_1}
\end{table}

\begin{table}[!ht]
\centering
\resizebox{\textwidth}{!}{%
\begin{tabular}{cccccccc}
\Xhline{1pt}
{\textbf{Template}} & {\textbf{Fine-grained}} & {\textbf{Demonstration}} & {\textbf{Aggregation}} & \textbf{Orca-7B} & \textbf{Orca-13B} & \textbf{Hermes-13B} & \textbf{Platypus-70B} \\ \hline\hline
Human                    & o                     & x                      & Approximation        & 0.3239          & 0.4002           & 0.2127             & 0.4041                  \\
Human                    & o                     & x                      & Logprob              & 0.3296          & 0.4210            & 0.2060              & 0.4305               \\
Human                    & o                     & x                      & Direct               & 0.3472          & 0.4468           & 0.2063             & 0.4354               \\
\rowcolor[HTML]{EFEFEF}
Model                    & o                     & x                      & Approximation        & 0.2687          & 0.3530            & 0.2152             & 0.4058                  \\
\rowcolor[HTML]{EFEFEF}
Model                    & o                     & x                      & Logprob              & 0.2926          & 0.3851           & 0.2250              & 0.4316                  \\
\rowcolor[HTML]{EFEFEF}
Model                    & o                     & x                      & Direct               & 0.2864          & 0.3844           & 0.2170              & 0.4039               \\ \Xhline{1pt}
\end{tabular}%
}
\caption{Comparison of Kendall's Tau correlation across various Score Aggregation and Models. \textit{Fine-grained} denotes whether the fine-grained scoring is used or not. \textit{Aggregation} denotes the type of Score Aggregation method used.}
\label{tab:tab_b_2}
\end{table}

\subsection{Comparisons of each Component}
Task description, evaluation criteria and evaluation steps of the prompt templates are slightly modified to ensure the suitability for each experiment. 
Examples are provided in Appendix~\ref{appendix:D}.

\subsubsection{Task Description type}
We investigate the impact of varying the length of task descriptions within the HG prompt and MG prompt on performance. 
Additionally, we compare performance when an expert role is assigned in the task description versus when it is not.
Various task descriptions are manually crafted for each prompt template, and examples can be found in Appendix~\ref{appendix:D}. 
The experimental results for the task description types can be found in Table~\ref{tab:tab_c_1}.


\begin{table}[!ht]
\centering
\resizebox{0.5\textwidth}{!}{%
\begin{tabular}{cccc}
\Xhline{1pt}
\textbf{Template}      & \textbf{Task Description} & \textbf{Orca-7B} & \textbf{Orca-13B} \\ \hline\hline
\multirow{4}{*}{Human} & Base                      & 0.3472           & 0.4468            \\
                       & Expert               & 0.3544  & 0.4383            \\
                       & Short                     & 0.3339           & 0.4239            \\
                       & Long                      & 0.3383           & 0.4501   \\ \hline
\multirow{4}{*}{Model} & Base                      & 0.2864           & 0.3744            \\
                       & Expert               & 0.3302  & 0.3881            \\
                       & Short                     & 0.2721           & 0.3508            \\
                       & Long                      & 0.2767           & 0.3891   \\ \Xhline{1pt}
\end{tabular}%
}
\caption{Comparison of Kendall's Tau correlation of cases using various types of task description on development set. Direct aggregation and fine-grained scoring are used for the experiment. Any demonstration is not provided.}
\label{tab:tab_c_1}
\end{table}

\subsubsection{Evaluation Criteria variants}
AD-GPT, HT-GPT, and MT-GPT are generated using GPT-4, tailored respectively to the AD, HT, and MT styles. 
The experimental results based on the types of the evaluation criteria can be found in Table~\ref{tab:tab_c_2}.

\begin{table}[!ht]
\centering
\resizebox{0.5\textwidth}{!}{%
\begin{tabular}{cccc}
\Xhline{1pt}
\textbf{Template}       & \textbf{Evaluation Criteria}   & \textbf{Orca-7B}               & \textbf{Orca-13B}              \\ \hline\hline
                        & AD                             & 0.3343                         & 0.4279                         \\
                        & AD-GPT                         & 0.3345                         & 0.4336                         \\
                        & \cellcolor[HTML]{FFFFFF}HT     & \cellcolor[HTML]{FFFFFF}0.3256 & \cellcolor[HTML]{FFFFFF}0.4192 \\
                        & \cellcolor[HTML]{FFFFFF}HT-GPT & \cellcolor[HTML]{FFFFFF}0.3293 & \cellcolor[HTML]{FFFFFF}0.4192 \\
                        & \cellcolor[HTML]{FFFFFF}MT     & \cellcolor[HTML]{FFFFFF}0.3303 & \cellcolor[HTML]{FFFFFF}0.4314 \\
\multirow{-6}{*}{Human} & \cellcolor[HTML]{FFFFFF}MT-GPT  & \cellcolor[HTML]{FFFFFF}0.3344 & \cellcolor[HTML]{FFFFFF}0.4297 \\ \hline
                        & AD                             & 0.3116                         & 0.4001                         \\
                        & AD-GPT                         & 0.3115                         & 0.4066                         \\
                        & \cellcolor[HTML]{FFFFFF}HT     & \cellcolor[HTML]{FFFFFF}0.3013 & \cellcolor[HTML]{FFFFFF}0.3904 \\
                        & \cellcolor[HTML]{FFFFFF}HT-GPT & \cellcolor[HTML]{FFFFFF}0.2987 & \cellcolor[HTML]{FFFFFF}0.3894 \\
                        & \cellcolor[HTML]{FFFFFF}MT     & \cellcolor[HTML]{FFFFFF}0.3141 & \cellcolor[HTML]{FFFFFF}0.4102 \\
\multirow{-6}{*}{Model} & \cellcolor[HTML]{FFFFFF}MT-GPT  & \cellcolor[HTML]{FFFFFF}0.3037 & \cellcolor[HTML]{FFFFFF}0.3949 \\ \Xhline{1pt}
\end{tabular}%
}
\caption{Comparison of Kendall's Tau correlation of cases using various types of evaluation criteria on development set. AD-GPT, HT-GPT, and MT-GPT denote AD, HT, and MT generated by GPT-4. Direct aggregation and fine-grained scoring are used for the experiment. Any demonstrated example is not provided.}
\label{tab:tab_c_2}
\end{table}

\subsubsection{Complexity of evaluation steps}
Complex evaluation steps are crafted using GPT-4 for both HG and MG prompt. 
Examples are provided in Appendix~\ref{appendix:D}. 
The experimental results for the evaluation steps can be found in Table~\ref{tab:tab_c_3}.


\begin{table}[!ht]
\centering
\resizebox{0.5\textwidth}{!}{%
\begin{tabular}{cccc}
\Xhline{1pt}
\textbf{Template}      & \textbf{Evaluation Steps} & \textbf{Orca-7B} & \textbf{Orca-13B} \\ \hline\hline
\multirow{2}{*}{Human} & Base                     & 0.3317           & 0.4135            \\
                       & Complex                  & 0.2969           & 0.4027            \\ \hline
\multirow{2}{*}{Model} & Base                     & 0.2866           & 0.3767            \\
                       & Complex                  & 0.2840           & 0.3751            \\ \Xhline{1pt}
\end{tabular}%
}

\caption{Comparison of Kendall's Tau correlation of base and complex evaluation steps on development set. Direct aggregation and fine-grained scoring are used for the experiment. No demonstrated example is provided to either method.}
\label{tab:tab_c_3}
\end{table}

\subsection{Error Analysis}

\begin{table}[!ht]
\centering
\resizebox{0.45\textwidth}{!}{%
\begin{tabular}{cccc}
\Xhline{1pt}
\multicolumn{1}{l}{} & \textbf{Error Type}      & \textbf{Base} & \textbf{Reason-best} \\ \hline\hline
0                    & Good                     & 50\%          & 69\%                 \\ \Xhline{0.5pt}
1                    & Inconsistent                & 11\%          & 17\%                 \\
2                    & Hallucination              & 36\%          & 6\%                  \\
3                    & Different Aspect & 6\%           & 8\%                  \\ \Xhline{1pt}
\end{tabular}%
}
\caption{Error Occurrence Ratio when RG prompt with and without `Reason-best' demonstration are used. In this analysis, we use Orca-13B to generate a score and rationale for each aspect. 
Error Type 1 means that the rationale is inconsistent with the score. 
Error Type 2 means that the rationale includes hallucinated information not mentioned in the source text and/or summary. 
Error Type 3 means that the rationale is about different aspect rather than the designated aspect.}

\label{tab:appendix_a_1}
\end{table}

\begin{table}[!ht]
\centering
\resizebox{\textwidth}{!}{%
\begin{tabular}{|cl}
\hline
\multicolumn{2}{|c}{\cellcolor[HTML]{EFEFEF}\textbf{Example}}                                                                                                                                                          \\ \hline
\multicolumn{1}{|c|}{\textbf{Source}}                                                               & \multicolumn{1}{l|}{\begin{tabular}[c]{@{}l@{}}Esteban Cambiasso has won all the major European competitions a player can during his illustrious career \\ but revealed that keeping Leicester City in the Premier League would be up there with the best. \\ The Foxes are currently seven points adrift at the bottom of the table, with only eight games remaining, \\ knowing that time is running out to save themselves. Cambiasso refuses to give up and admits that keeping \\ Leicester up will feel like winning a trophy. Esteban Cambiasso says that helping keep Leicester in the \\ Premier League will feel like winning a trophy `For me, it's like another cup,' he told BBC East Midlands Today. \\ `When you start another season you have an objective, and this is the objective for us. `For me, winning a cup \\ or winning the league with another team is the same now as having the possibility to save Leicester in the \\ Premier League.' The Argentinian midfielder poses with the trophy after his team won the 2010 FIFA Club \\ World Cup Cambiasso had an illustrious career at Inter Milan, winning an impressive 15 trophies during his \\ stint River Plate (2001-2002) Argentine Primera Division Real Madrid (2002-2004) La Liga Super Cup \\ Supercopa de Espana Inter Milan (2004-2014) Champions League Serie A (5) Coppa Italia (4) Supercoppa \\ (4) FIFA Club World Cup Having not won a game since January, Nigel Pearson's men face West Ham United \\ on Saturday and Cambiasso is still convinced they can avoid the drop. `I understood when I signed for Leicester \\ it's not an easy job to stay in the Premier League,' he said. `It's a difficult situation but I think we have our \\ chances to win matches. There's a quarter of the Premier League left to finish. `I think some people think for \\ Leicester all is finished. But I'm sure, because I watch my team-mates every day, we can save Leicester and stay \\ in the Premier League.' The former Inter Milan star signed for the Foxes in the summer, leaving Italy after ten \\ years and embarking on a new challenge in England. After agreeing to a one-year-deal, Cambiasso has quickly \\ established himself as a key player but it remains to be seen if he'll still be in the East Midlands at the start \\ of next season. The former Real Madrid man was also successful during his short spell in Spain for Real Madrid \\ Cambiasso played during Real's `Galatico' era, with Luis Figo, Zinedine Zidane, Ronaldo and David Beckham \\ `Leicester always wanted me,' he added. `After these nine months or eight months, I'm very happy because \\ my family is OK, and I'm good. `I want a few more points, but all the rest is perfect.' Cambiasso is happy in \\ the East Midlands and could stay beyond his current one-year-deal\end{tabular}} \\ \hline
\multicolumn{1}{|c|}{\textbf{Summary}}                                                              & \multicolumn{1}{l|}{Esteban Cambiasso is confident Leicester can win the Premier League and keep themselves in the top flight.}                                                                                                                                        \\ \hline
\multicolumn{1}{|c|}{\textbf{Score}}                                                                & \multicolumn{1}{l|}{1}                                                    \\ \hline
\multicolumn{1}{|c|}{\textbf{\begin{tabular}[c]{@{}c@{}}Rationale\end{tabular}}} & \multicolumn{1}{l|}{\begin{tabular}[c]{@{}l@{}}\textbf{The summary is coherent with the source text}, as it captures the main points about Esteban Cambiasso's confidence \\ in Leicester's chances of staying in the Premier League. However, it could be improved by including more details \\ and quotes from the source text, making it more comprehensive and engaging.\end{tabular}}                                                                                                                                                                                                                                           \\ \hline
\end{tabular}%
}

\caption{Example of Error Type 1. The rationale is inconsistent with the score for the \texttt{coherence}.}
\label{tab:appendix_a_2_1}
\end{table}

\clearpage
\begin{table}[!ht]
\centering
\resizebox{\textwidth}{!}{%
\begin{tabular}{|cl}
\hline
\multicolumn{2}{|c}{\cellcolor[HTML]{EFEFEF}\textbf{Example}}                                                                                                                                                                                                                  \\ \hline
\multicolumn{1}{|c|}{\textbf{Source}}                                                               & \multicolumn{1}{l|}{\begin{tabular}[c]{@{}l@{}}"(CNN)Two passengers found dead on a cruise ship in Puerto Rico appear to have died in a murder-suicide, the cruise line said.\\ Holland America Line said two guests were found dead inside their stateroom on the ms Ryndam at 11:30 a.m. Thursday. "\\ "The cabin was immediately secured, and the authorities were notified, including the FBI,"" Holland America said.\\ ""We are cooperating fully with the investigation, and the authorities will make the official determination on what occurred.""\\ FBI spokesman Moises Quiñones said authorities were on scene investigating. The ship left Tampa, Florida, on March 29\\ on a 14-day Southern Caribbean cruise. It’s currently in San Juan, Puerto Rico. Puerto Rico Port Authority spokesman\\ Efraín Santiago told El Nuevo Dia newspaper that the cleaning staff on the ship had discovered the deceased passengers\\ after knocking on the cabin's door.\end{tabular}} \\ \hline
\multicolumn{1}{|c|}{\textbf{Summary}}                                                              & \multicolumn{1}{l|}{\begin{tabular}[c]{@{}l@{}}holland america line said two guests were found dead inside their stateroom on the ship .\\ the ship left tampa , florida , on march 29 on a 14-day southern caribbean cruise . it's currently in san juan , puerto rico\end{tabular}}\\ \hline
\multicolumn{1}{|c|}{\textbf{Score}}                                                                & \multicolumn{1}{l|}{2}    \\ \hline
\multicolumn{1}{|c|}{\textbf{\begin{tabular}[c]{@{}c@{}}Rationale\end{tabular}}} & \multicolumn{1}{l|}{\begin{tabular}[c]{@{}l@{}}The summary is not coherent with the source text. It omits crucial information, such as the cause of death,\\ the identities of the passengers, and the cruise ship’s itinerary. The summary also does not convey the urgency and\\ seriousness of the situation, as indicated by the FBI’s involvement and \textbf{the ship’s immediate return to port}. The coherence\\ of the summary is poor, as it does not represent the main events and context of the source text\end{tabular}}                                        \\ \hline
\end{tabular}%
}
\caption{Example of Error Type 2. Rationale includes hallucination for \texttt{coherence}.}
\label{tab:appendix_a_2_2}
\end{table}

\begin{table}[!ht]
\centering
\resizebox{\textwidth}{!}{%
\begin{tabular}{|cl}
\hline
\multicolumn{2}{|c}{\cellcolor[HTML]{EFEFEF}\textbf{Example}}                                                                             \\ \hline
\multicolumn{1}{|c|}{\textbf{Source}}                                                               & \multicolumn{1}{l|}{\begin{tabular}[c]{@{}l@{}}First blood to United after their Under 18s saw off City 1-0 in the `mini-derby'. Kits aside, this was probably as far removed \\ from Sunday's big match as you could imagine. For a start, no less than 13 out of the 22 players on show at kick-off were English. \\ Of those, 10 were Mancunian. Callum Gribbin was the matchwinner for Manchester United with a delightful free-kick Ticket prices? \\ Entry was free and close to 1,000 gathered on the seats and terraces of Moss Lane for a match that kicked off at 3pm on Saturday \\ with half-and-half scarves nowhere to be seen. Altrincham's compact 6,000-capacity home may not be Old Trafford, but it does \\ have a proud history of its own. It was certainly a grander stage than a windswept Saturday morning on an outfield at United's \\ Carrington complex, where Paul McGuinness's Under 18s usually ply their trade. The young Reds coach wanted to make the \\ experience closer to what his tyros could expect should they make the step up to the seniors. And his side lined up with three \\ at the back and supporting wingbacks in a formation seen more than once this season in the first team. In an even first-half \\ the impressive Marcus Wood, from just down the road in Sale, came closest for City with an audacious chip. United manager \\ Paul McGuinness saw his side claim victory in the `mini derby' For the home side towering centre-forward Marcus Rashford, \\ another local lad from whom big things are expected, wasted two decent opportunities when put through. Just before the \\ hour mark England Under 17 star Callum Gribbin made and scored the first. In March, the nifty-footed attacking midfielder \\ went viral when United tweeted a clip of him showing outrageous skills to trick four Aston Villa players. He was at it again \\ here, showing nifty footwork on the edge of the box and drawing a foul. After dusting himself down Gribbin stepped up to curl \\ a delightful free-kick around the wall and beyond blonde Joe Hart-a-like Charlie Albinson in the City goal. Moments later it was \\ almost two when full-back Cameron Borthwick-Jackson's rasping 25-yard drive struck the angle of the crossbar and post. The same \\ man looked to have gotten away with one when he appeared to clip Buckley in the box before sub Aaron Nemane hit the side-netting \\ as the visitors looked for an equaliser. Dutch winger Javairo Dilrosun curled a free kick onto the roof of the net in stoppage time for \\ City but that was all she wrote. The result leaves United, who have won two of this year's cross-town clashes, two points behind l\\ eaders Middlesbrough with a game in hand. Injury-ravaged City however, have now lost both matches in the Barclays Under 18s \\ Premier League end-of-season play-off series. FA chief Greg Dyke recently bemoaned the lack of first-team opportunities for homegrown \\ talent between the ages of 18 and 21. On the evidence of this entertaining 90 minutes high on skill and commitment, there are talents \\ on both sides that may well deserve their chance in the On the evidence of this entertaining 90 minutes high on skill and commitment, \\ there are talents on both sides that may well deserve their chance in the coming years.\end{tabular}} \\ \hline
\multicolumn{1}{|c|}{\textbf{Summary}}                                                              & \multicolumn{1}{l|}{\begin{tabular}[c]{@{}l@{}}callum gribbin was the matchwinner for manchester united with a free-kick . paul mcguinness's side claim victory in the \\ `mini derby' for a match. paul mcguinness and marcus wood were in the first team .\end{tabular}}                                                                                                                                                                                    \\ \hline
\multicolumn{1}{|c|}{\textbf{Score}}                                                                & \multicolumn{1}{l|}{3}                                                                                                                                                                            \\ \hline
\multicolumn{1}{|c|}{\textbf{\begin{tabular}[c]{@{}c@{}}Rationale\end{tabular}}} & \multicolumn{1}{l|}{\begin{tabular}[c]{@{}l@{}}The summary is not entirely consistent with the source document. It does mention Callum Gribbin as the matchwinner and the victory \\ in the `mini derby', but it omits important details such as the final score, the formation used by Manchester United, and the \\ performance of Marcus Wood.\end{tabular}}                                                                                                      \\ \hline
\end{tabular}%
}
\caption{Example of Error Type 3. The rationale does not discuss for \texttt{consistency}.}
\label{tab:appendix_a_2_3}
\end{table}

\section{Example Prompts}
\label{appendix:D}
\begin{table}[!ht]
\resizebox{0.98\textwidth}{!}{%
\begin{tabular}{|c|c|l|}
\hline
\rowcolor[HTML]{EFEFEF}
\textbf{Task Description} & \textbf{Template} & \multicolumn{1}{c|}{\textbf{Prompt}} \\ \hline
Expert & Human & \begin{tabular}[c]{@{}l@{}}You read and summarize a lot of news articles, and you're an expert at summarizing news articles. \\ In this task you will evaluate the quality of a summary written for a news article. \\ To correctly solve this task, follow these steps:\end{tabular} \\ \hline
Expert & Model & \begin{tabular}[c]{@{}l@{}}You read and summarize a lot of news articles, and you're an expert at summarizing news articles. \\ You will be given one summary written for a news article. Your task is to evaluate the summary \\ based on a specific metric, rating it on a scale from 1 (worst) to 5 (best). \\ Please make sure you read and understand these instructions carefully. \\ Please keep this document open while reviewing, and refer to it as needed.\end{tabular} \\ \hline
Long & Human & \begin{tabular}[c]{@{}l@{}}In this task, you will evaluate the quality of a summary written for a news article. \\ Please take your time to carefully evaluate the provided summary, and don't hesitate to refer back \\ to this instruction document if you need clarification or guidance at any point during your evaluation. \\ To correctly solve this task, follow these steps:\end{tabular} \\ \hline
Long & Model & \begin{tabular}[c]{@{}l@{}}You will be given one summary written for a news article.\\ Your task is to evaluate the summary based on a specific metric, rating it on a scale from 1 (worst) \\ to 5 (best). Please make sure you read and understand these instructions carefully. \\ Please keep this document open while reviewing, and refer to it as needed.Please take your time \\ to carefully evaluate the provided summary, and don't hesitate to refer back to this instruction document \\ if you need clarification or guidance at any point during your evaluation.\end{tabular} \\ \hline
Short & Human & Evaluate the news article summary quality. \\ \hline
Short & Model & \begin{tabular}[c]{@{}l@{}}Evaluate a news article summary using a specific metric, rating it from 1 (worst) to 5 (best).\\ Please read and understand these instructions carefully. Keep this document open for reference \\ while reviewing.\end{tabular} \\ \hline
\end{tabular}%
}
\onecolumn\caption{Examples of different variants of Task Description}
\label{tab:appendix_d_1}
\end{table}

\begin{table}[!ht]
\resizebox{0.98\textwidth}{!}{%
\begin{tabular}{|c|c|l|}
\hline
\rowcolor[HTML]{EFEFEF}
\textbf{Evaluation Criteria} & \textbf{Template} & \multicolumn{1}{c|}{\textbf{Prompt}} \\ \hline
HT-GPT & Human & \begin{tabular}[c]{@{}l@{}}Relevance:This rating assesses the extent to which the summary highlights the central themes \\ of the original article. Evaluate if the summary encompasses the crucial elements while omitting \\ any non-essential details.\end{tabular} \\ \hline
MT-GPT & Model & \begin{tabular}[c]{@{}l@{}}Relevance - gauges the summary's alignment with the article's primary ideas. Check if the \\ summary includes essential points and omits unrelated details. It may help to list the article's \\ main points and verify their presence in the summary.\end{tabular} \\ \hline
AD & Human,Model & Relevance - How well is the generated text relevant to its source text? \\ \hline
AD-GPT & Human,Model & Relevance - To what extent does the generated summary capture and reflect the core details of its source text? \\ \hline
\end{tabular}%
}
\onecolumn\caption{Examples of different variants of Evaluation Criteria}
\label{tab:appendix_d_2}
\end{table}

\begin{table}[!ht]
\centering
\resizebox{0.98\textwidth}{!}{%
\begin{tabular}{|c|c|l|}
\hline
\rowcolor[HTML]{EFEFEF} 
\textbf{Evaluation Steps} &
  \textbf{Template} &
  \multicolumn{1}{c|}{\cellcolor[HTML]{EFEFEF}\textbf{Prompt}} \\ \hline
 &
  Human &
  \begin{tabular}[c]{@{}l@{}}\\In this task, your primary aim is to conduct a thorough assessment of the summary provided for a news article.\\ To effectively accomplish this task, please adhere to the following comprehensive steps:\\ \\ 1. Initiate the evaluation process by engaging in an in-depth examination of the news article.\\ Your aim here is to establish a profound understanding of the article’s entire spectrum of content,\\ ensuring you grasp its core message, nuances, and key elements.\\ \\ 2. Proceed to scrutinize the proposed summary provided alongside the article.\\ In this phase, your task is to meticulously evaluate the summary for its aspect.\\ \\ 3. Assign a rating to each summary based on its aspect,\\ utilizing a scale ranging from 1 (indicating the lowest quality) to 5 (signifying the highest quality).\end{tabular} \\ \cline{2-3} 
\multirow{-15}{*}{Complex} &
  Model &
  \begin{tabular}[c]{@{}l@{}}\\1. Thoroughly examine the provided summary and the source document with meticulous attention to detail.\\ \\ 2. Conduct a comprehensive comparative analysis, scrutinizing the summary in relation to the source document\\ to discern and delineate the primary focal points and pivotal elements elucidated within the article.\\ \\ 3. Engage in a judicious evaluation to gauge the summary’s efficacy\\ in addressing and encompassing the central facets of the source document,\\ concurrently assessing the presence of any extraneous or duplicative information that might detract from its relevance.\\ \\ 4. Utilize a relevance rating scale, ranging from 1 (indicating minimal relevance) to 5 (indicating maximal relevance),\\ for the purpose of assigning a numerical score.\\ This score serves as a quantitative reflection of the extent to which the summary aligns with\\ and encapsulates the core substance of the source document.\end{tabular} \\ \hline
\end{tabular}%
}
\caption{Examples of Complex Evaluation Steps}
\label{tab:complex}
\end{table}

\begin{table}[!ht]
\resizebox{\textwidth}{!}{%
\begin{tabular}{|c|c|l|}
\hline
\cellcolor[HTML]{EFEFEF}\textbf{Template} & \multicolumn{1}{c|}{\cellcolor[HTML]{EFEFEF}\textbf{Prompt}} \\ \hline
 Human, Model, Rationale &  \begin{tabular}[c]{@{}l@{}}Please refer to following example below.\\ Source text: Twice French Open champion Serena Williams said her struggle to beat Sara Errani i\\ n the Fed Cup on Sunday had been a real `eye-opener ' as the claycourt season gets into full swing . \\ World No 1 Williams eventually prevailed 4-6 7-6 ( 3 ) 6-3 against the dogged Italian to take her career \\ record over her to 8-0 but the American was not impressed . The US were beaten 3-2 as Williams \\ and Alison Riske were thrashed 6-0 6-3 in the doubles rubber by Errani and Flavia Pennetta , \\ meaning they were relegated to World Group II . American tennis star Serena Williams fought back \\ to beat Italian Sara Errani in the Fed Cup play-off on Sunday Tough weather conditions made it \\ difficult for both players who had to keep on re-tossing their serves Errani gave Williams a real scare \\ but in the end the world No 1 's power proved to be too much `Today has been a big eye opener , \\ ' Williams said afterwards . ` I 'm totally not as ready for the claycourt season as I thought I was . \\ Now I 'm in the mindset of , `` You know what , I 'm not on hard court . `` I 'm playing like I 'm on hard \\ court and I 'm not . `So I have to play and be ready to hit a thousand shots if necessary . ' Williams , 33 , \\ won her 19th singles grand slam at the Australian Open and her dominance has raised talk of her \\ claiming all the majors this year . The French Open has been her least successful of the four though \\ despite claiming the title in Paris in 2002 and 2013 . Her doubles defeat on Sunday blotted an otherwise \\ flawless Fed Cup record and left the US facing a battle to get back amongst the elite nations next year . \\ `We have to work harder , ' US captain Mary Joe Fernandez said . `We came close today and need to \\ just keep plugging away . 'The good news is that we have a lot of players in the top 100 and , hopefully , \\ we can get two wins next year and get back into the World Group . ` Williams congratulates Italy captain \\ Corrado Barazzutti after competing in America 's doubles defeat.\\ Summary: Serena Williams beat Sara Errani 4-6 7-6 ( 3 ) 6-3 in the Fed Cup play-off . \\ The US were beaten 3-2 as Williams and Alison Riske were thrashed in the doubles rubber . \\ The doubles defeat saw the US relegated to World Group II .\textbackslash{}u2019\\ \\ \-----\\ Example Score: 5\\ Explanation: The summary effectively captures the key points from the article. It mentions Serena \\ Williams' challenging match against Sara Errani and her eventual victory. The summary also highlights \\ the US team's overall defeat and its consequence \textbackslash{}u2013 relegation to World Group II. These details \\ are central to the main storyline of the source text, making the summary highly relevant. Thus, a score \\ of 5 (best) is appropriate for the summary's relevance.\end{tabular} \\ \hline
\end{tabular}%
}
\onecolumn\caption{Example of Demonstration with rationale}
\label{tab:appendix_d_3}
\end{table}

\begin{table}[!ht]
\resizebox{\textwidth}{!}{%
\begin{tabular}{|c|c|l|}
\hline
\cellcolor[HTML]{EFEFEF}\textbf{Template} & \multicolumn{1}{c|}{\cellcolor[HTML]{EFEFEF}\textbf{Prompt}} \\ \hline
Rationale & \begin{tabular}[c]{@{}l@{}}Your task is to evaluate the relevance of a provided summary based on its source document.\\ Follow these steps:\\ \\ 1. Read the source document\\ 2. Review the summary\\ 3. Analyze for relevance\\ 4. Assign a Score: Rate the summary on a scale of 1 to 5, where:\\ - 1 means the summary is not relevant with the source.\\ - 5 means the summary is entirely relevant with the source.\\ 5. Provide a Rationale: After assigning a score, explain your reasons based on your analysis.\\ \\ \# Definition:\\Relevance:\\The rating measures how well the summary captures the key points of the article.\\Consider whether all and only the important aspects are contained in the summary."
\\ -----\\ Source text: \\ Summary: \end{tabular} \\ \hline
\end{tabular}%
}

\onecolumn\caption{Example of Rationale Generation(RG) prompt}
\label{tab:appendix_d_5}
\end{table}

\begin{table}[!ht]
\resizebox{\textwidth}{!}{%
\begin{tabular}{|l|l|l|}
\hline
\multicolumn{1}{|c|}{\cellcolor[HTML]{EFEFEF}\textbf{Template}} & \multicolumn{1}{c|}{\cellcolor[HTML]{EFEFEF}\textbf{Prompt}} \\ \hline
Filtering & \begin{tabular}[c]{@{}l@{}}In this task you will evaluate the quality of a summary written for a document.\\ \\ Provided summary may include direct or rephrased repetitions of the same word or phrase. \\ \\ With that in mind do the following:\\ \\ 1. Answer whether the summary is redundant or not.\\ - Your answer must be in "Yes" or "No" format, where "Yes" means that the summary is redundant and \\ "No" means that the summary is not redundant.\\ \\ 2. Please provide brief explanation for your answer.\\ - Your explanation should only discuss the redundancy of the summary, not the quality of the summary \\ in general.\\ ----\\ summary: \end{tabular} \\ \hline
\end{tabular}%
}
\onecolumn\caption{Example of Filtering prompt}
\label{tab:appendix_d_6}
\end{table}

\begin{table}[!ht]
\centering
\resizebox{\textwidth}{!}{%
\begin{tabular}{|c|l|}
\hline
\rowcolor[HTML]{EFEFEF} 
\textbf{Template} & \multicolumn{1}{c|}{\cellcolor[HTML]{EFEFEF}\textbf{Prompt}}                                                                                                                                                                                                                                                                                                                                                                                                                                                                                                                                                                       \\ \hline
Baseline             & \begin{tabular}[c]{@{}l@{}}Score the summarization with respect to the summarized document on a continuous scale from 0 to 100, \\ where a score of zero means irrelevant, factually incorrect and not readable and score of one hundred means, \\relevant, factually correct, good readability\\ \\ ----\\ Source text: \\ \\ Summary: \end{tabular} \\ \hline
\end{tabular}%
}
\caption{Example of Baseline prompt}
\label{tab:baseline_full_prompt}
\end{table}

\begin{table}[!ht]
\centering
\resizebox{\textwidth}{!}{%
\begin{tabular}{|c|c|l|}
\hline
\rowcolor[HTML]{EFEFEF} 
\textbf{Template} & \multicolumn{1}{c|}{\cellcolor[HTML]{EFEFEF}\textbf{Prompt}}                                                                                                                                                                                   \\ \hline
Model             & \begin{tabular}[c]{@{}l@{}}You will be given one summary written for a news article.\\ Your task is to rate the summary on one metric.\\ Please make sure you read and understand these instructions carefully. \\ Please keep this document open while reviewing, and refer to it as needed.\\ \\ Evaluation Criteria:\\ Relevance - selection of important content from the source. The summary should \\ include only important information from the source document. Annotators were\\ instructed to penalize summaries which contained redundancies and excess information.\\ \\ \\ Evaluation Steps:\\ 1. Read the summary and the source document carefully.\\ 2. Compare the summary to the source document and identify the main points of the article.\\ 3. Assess how well the summary covers the main points of the article, and how much irrelevant \\ or redundant information it contains.\\ 4. Assign a relevance score from 1 to 5.\\ \\ Example:\\ Source Text:\\ \\ Summary:\\ \\ Evaluation Form (scores ONLY):\\ - Relevance:\end{tabular} \\ \hline
\end{tabular}%
}
\caption{Example of Model Guideline(MG) prompt}
\label{tab:model_full_prompt}
\end{table}

\begin{table}[!ht]
\centering
\resizebox{\textwidth}{!}{%
\begin{tabular}{|c|l|}
\hline
\rowcolor[HTML]{EFEFEF} 
\textbf{Template} & \multicolumn{1}{c|}{\cellcolor[HTML]{EFEFEF}\textbf{Prompt}}                                                                                                                                                                                                                                                                                                                                                                                                                                                                                                                                                                       \\ \hline
Human             & \begin{tabular}[c]{@{}l@{}}In this task you will evaluate the quality of a summary written for a document.\\ \\ To correctly solve this task, follow these steps:\\ \\ 1. Carefully read the document, be aware of the information it contains.\\ 2. Read the proposed summary.\\ 3. Rate each summary on a scale from 1 (worst) to 5 (best) by its relevance.\\ \\ \# Definition:\\Relevance: The rating measures how well the summary captures the key points of the article.\\ Consider whether all and only the important aspects are contained in the summary.\\ Source text: \\ \\ Summary: \\ \\ Score:\end{tabular} \\ \hline
\end{tabular}%
}
\caption{Example of Human Guideline(HG) prompt}
\label{tab:human_full_prompt}
\end{table}

\begin{table}[!ht]
\centering
\resizebox{\textwidth}{!}{%
\begin{tabular}{|c|l|}
\hline
\rowcolor[HTML]{EFEFEF} 
\textbf{Template} &
  \multicolumn{1}{c|}{\cellcolor[HTML]{EFEFEF}\textbf{Prompt}} \\ \hline
Human &
  \begin{tabular}[c]{@{}l@{}}Instruction:\\ In this task you will evaluate the quality of a summary written for a document. \\ \\ To correctly solve this task, follow these steps:\\ \\ 1. Carefully read the document, be aware of the information it contains.    \\ 2. Read the proposed summary.\\ 3. Rate each summary on a scale from 0 (worst) to 100 (best) by its Relevance. \\ \\ \# Definition:\\ Relevance: The rating measures how well the summary captures the key points of the article.\\ Consider whether all and only the important aspects are contained in the summary.\\ Source text:\\ \\ \\ Summary:\\ \\ Score:\end{tabular} \\ \hline
\end{tabular}%
}
\caption{Example of Human Guideline(HG) prompt of \texttt{relevance} used in test phase}
\label{tab:test_relevance}
\end{table}

\begin{table}[]
\centering
\resizebox{\textwidth}{!}{%
\begin{tabular}{|c|l|}
\hline
\rowcolor[HTML]{EFEFEF} 
\textbf{Template} &
  \multicolumn{1}{c|}{\cellcolor[HTML]{EFEFEF}\textbf{Prompt}} \\ \hline
Human &
  \begin{tabular}[c]{@{}l@{}}Instruction:\\ In this task you will evaluate the quality of a summary written for a document. \\ \\ To correctly solve this task, follow these steps:\\ \\ 1. Carefully read the document, be aware of the information it contains.   \\ 2. Read the proposed summary.\\ 3. Rate each summary on a scale from 0 (worst) to 100 (best) by its Factuality. \\ \\ \# Definition:\\ Factuality: This rating gauges the accuracy and truthfulness of the information presented \\ in the summary compared to the original article.\\ Scrutinize the summary to ensure it presents facts without distortion or misrepresentation, \\ staying true to the source content's details and intent.\\ Source text: \\ \\ Summary: \\ \\ Score:\end{tabular} \\ \hline
\end{tabular}%
}
\caption{Example of Human Guideline(HG) prompt of \texttt{factuality} used in test phase}
\label{tab:test_factuality}
\end{table}

\begin{table}[]
\centering
\resizebox{\textwidth}{!}{%
\begin{tabular}{|c|l|}
\hline
\rowcolor[HTML]{EFEFEF} 
\textbf{Template} &
  \multicolumn{1}{c|}{\cellcolor[HTML]{EFEFEF}\textbf{Prompt}} \\ \hline
Human &
  \begin{tabular}[c]{@{}l@{}}Instruction:\\ In this task you will evaluate the quality of a summary written for a document. \\ \\ To correctly solve this task, follow these steps:\\ \\ 1. Carefully read the document, be aware of the information it contains.    \\ 2. Read the proposed summary. \\ 3. Rate each summary on a scale from 0 (worst) to 100 (best) by its Fluency. \\ \\ \# Definition:\\ Fluency: This rating evaluates the clarity and grammatical integrity of each sentence in the summary.\\ Examine each sentence for its structural soundness and linguistic clarity.\\ Source text: \\ \\ \\ Summary: \\ \\ Score:\end{tabular} \\ \hline
\end{tabular}%
}
\caption{Example of Human Guideline (HG) prompt of \texttt{fluency} used in test phase}
\label{tab:test_fluency}
\end{table}

\end{document}